\def\year{2022}\relax
\documentclass[letterpaper]{article} % DO NOT CHANGE THIS
\usepackage{aaai22}  % DO NOT CHANGE THIS
\usepackage{times}  % DO NOT CHANGE THIS
\usepackage{helvet}  % DO NOT CHANGE THIS
\usepackage{courier}  % DO NOT CHANGE THIS
\usepackage[hyphens]{url}  % DO NOT CHANGE THIS
\usepackage{graphicx} % DO NOT CHANGE THIS
\usepackage{natbib}  % DO NOT CHANGE THIS AND DO NOT ADD ANY OPTIONS TO IT
\usepackage{caption} % DO NOT CHANGE THIS AND DO NOT ADD ANY OPTIONS TO IT
\DeclareCaptionStyle{ruled}{labelfont=normalfont,labelsep=colon,strut=off} % DO NOT CHANGE THIS
\usepackage{algorithm}
\usepackage{algorithmic}

\usepackage{newfloat}
\usepackage{listings}
\floatstyle{ruled}
\newfloat{listing}{tb}{lst}{}
\floatname{listing}{Listing}

\usepackage{times}
\usepackage{latexsym}

\usepackage{microtype}
\usepackage{adjustbox}
\usepackage{subfigure}
\usepackage{amsmath}
\usepackage{booktabs} % for professional tables

\usepackage{amssymb}
\usepackage{multirow}
\usepackage{makecell}
\usepackage{xcolor}
\usepackage{seqsplit}
\newcommand{\alert}[1]{\textcolor{red}{#1}}
\newcommand{\bertlarge}{\ensuremath{\text{BERT}_{\textsc{LARGE }}}}
\newcommand{\bartlarge}{\ensuremath{\text{BART}_{\textsc{LARGE }}}}
\newcommand{\tfivebase}{\ensuremath{\text{T5}_{\textsc{BASE }}}}
\newcommand{\robertabase}{\ensuremath{\text{RoBERTa}_{\textsc{BASE }}}}
\newcommand{\robertalarge}{\ensuremath{\text{RoBERTa}_{\textsc{LARGE }}}}

\title{Semantic Parsing in Task-Oriented Dialog \\with Recursive Insertion-based Encoder}
\author{Elman Mansimov and Yi Zhang}
\affiliations{AWS AI Labs\\\{mansimov, yizhngn\}@amazon.com}
\usepackage{bibentry}
\begin{document}

\maketitle

\begin{abstract}
We introduce \textbf{R}ecursive \textbf{IN}sertion-based \textbf{E}ncoder (RINE), a novel approach for semantic parsing in task-oriented dialog. Our model consists of an encoder network that incrementally builds the semantic parse tree by predicting the non-terminal label and its positions in the linearized tree. At the generation time, the model constructs the semantic parse tree by recursively inserting the predicted non-terminal labels at the predicted positions until termination. RINE achieves state-of-the-art exact match accuracy on low- and high-resource versions of the conversational semantic parsing benchmark {\tt TOP} \citep{Gupta2018SemanticPF,Chen2020LowResourceDA}, outperforming strong sequence-to-sequence models and transition-based parsers. We also show that our model design is applicable to nested named entity recognition task, where it performs on par with state-of-the-art approach designed for that task. Finally, we demonstrate that our approach is $2-3.5 \times$ faster than the sequence-to-sequence model at inference time.
\end{abstract}

\section{Introduction}
Task-oriented dialog systems are playing an increasingly important role in modern business and social lives of people by facilitating information access and automation of routine tasks through natural language conversations. At the core of such dialog systems, a natural language understanding component interprets user input utterances into a meaning representation. While the traditional intent-slot based approach can go a long way, such flat meaning representation falls short of capturing the nuances of natural languages, where phenomena such as conjunction, negation, co-reference, quantification and modification call for a hierarchically structured representation, as illustrated by recent work \cite{Gupta2018SemanticPF,bonial-etal-2020-dialogue,Cheng2020ConversationalSP,Andreas2020TaskOrientedDA}.
Commonly adopted tree- or directed acyclic graph-based structures resemble traditional frameworks for syntactic or semantic parsing of natural language sentences.

%Parsing the input sentence into the representation of its meaning is the core component of the task-oriented dialogue systems. Traditionally the meaning representation of the dialogue systems has been represented by the flat structure that contains \textit{intent} (action requested by the user) and \textit{slots} (entities corresponding to the specified action). Traditional rule-based or slot-filling system classifies the sentence with single intent class and tags each input token with the slot label \citep{Mesnil2013InvestigationOR,Liu2016AttentionBasedRN}. For example, for the sentence \textit{Find flights to New York} the slot-filling model would predict the {\tt GET\_AVAILABLE\_FLIGHTS} intent and tag tokens \textit{New York} as the {\tt DESTINATION} slot. 

%While the flat meaning representation can go a long way, it is difficult to represent compositional queries that contain nested intents and slots by such representation. To support such use-case, \citet{Gupta2018SemanticPF} proposed the hierarchical meaning representation for the dialogue systems that is similar to design of standard constituency parses. For instance, the query \textit{Driving directions to the Eagles game} is composed of {\tt GET\_DIRECTIONS} and {\tt GET\_EVENT} intents nested in the single tree in the hierarchical representation \citep{Gupta2018SemanticPF}.

The hierarchical representation of \citet{Gupta2018SemanticPF} motivated the extension of neural shift-reduce parsers \citep{Dyer2016RecurrentNN,Einolghozati2019ImprovingSP}, neural span-based parsers \citep{Stern2017AMS,Pasupat2019SpanbasedHS} and sequence-to-sequence (seq2seq) \citep{Sutskever2014SequenceTS, vaswani2017attention,Rongali2020DontPG} models for handing \textit{compositional} queries in task-oriented dialog. Due to the state-of-the-art performance of these models, there has been limited work designing structured prediction models that have a stronger inductive bias for semantic parsing of task-oriented dialog utterances.

In this paper, we propose the \textbf{R}ecursive \textbf{In}sertion-based \textbf{E}ncoder) (RINE; pronounced "Ryan"), that incrementally builds the semantic parse tree by inserting the non-terminal intent/slot labels into the utterance. The model is trained as a discriminative model that predicts labels with their corresponding positions in the input. At generation time, the model constructs the semantic parse tree by recursively inserting the predicted label at the predicting position until the termination. Unlike seq2seq models \citep{Rongali2020DontPG,Zhu2020DontPI,Aghajanyan2020ConversationalSP,Babu2021NonAutoregressiveSP}, our approach does not contain a separate decoder which generates the linearized semantic parse tree.%While we focus on semantic parsing in task-oriented dialog, our proposed approach is applicable to other parsing tasks such as constituency parsing and hierarchical named entity recognition.

We extensively evaluate our proposed approach on low-resource and high-resource versions of the popular conversational semantic parsing dataset {\tt TOP} \citep{Gupta2018SemanticPF,Chen2020LowResourceDA}. We compare our model against a state-of-the-art transition-based parser RNNG \citep{Gupta2018SemanticPF,Einolghozati2019ImprovingSP} and seq2seq models \citep{Rongali2020DontPG,Zhu2020DontPI,Aghajanyan2020ConversationalSP,Babu2021NonAutoregressiveSP} adapted to this task. We show that our approach achieves the state-of-the-art performance on both low-resource and high-resource settings {\tt TOP}. In particular, RINE achieves \textbf{up to an $13\%$ \iffalse $11\%$ \fi absolute improvement in exact match} in the low-resource 
setting. We also demonstrate that our approach is $2-3.5 \times$ faster than strong sequence-to-sequence model at inference time.

\begin{figure*}
    \centering
    \includegraphics[width=0.99\textwidth,height=0.99\textheight,keepaspectratio]{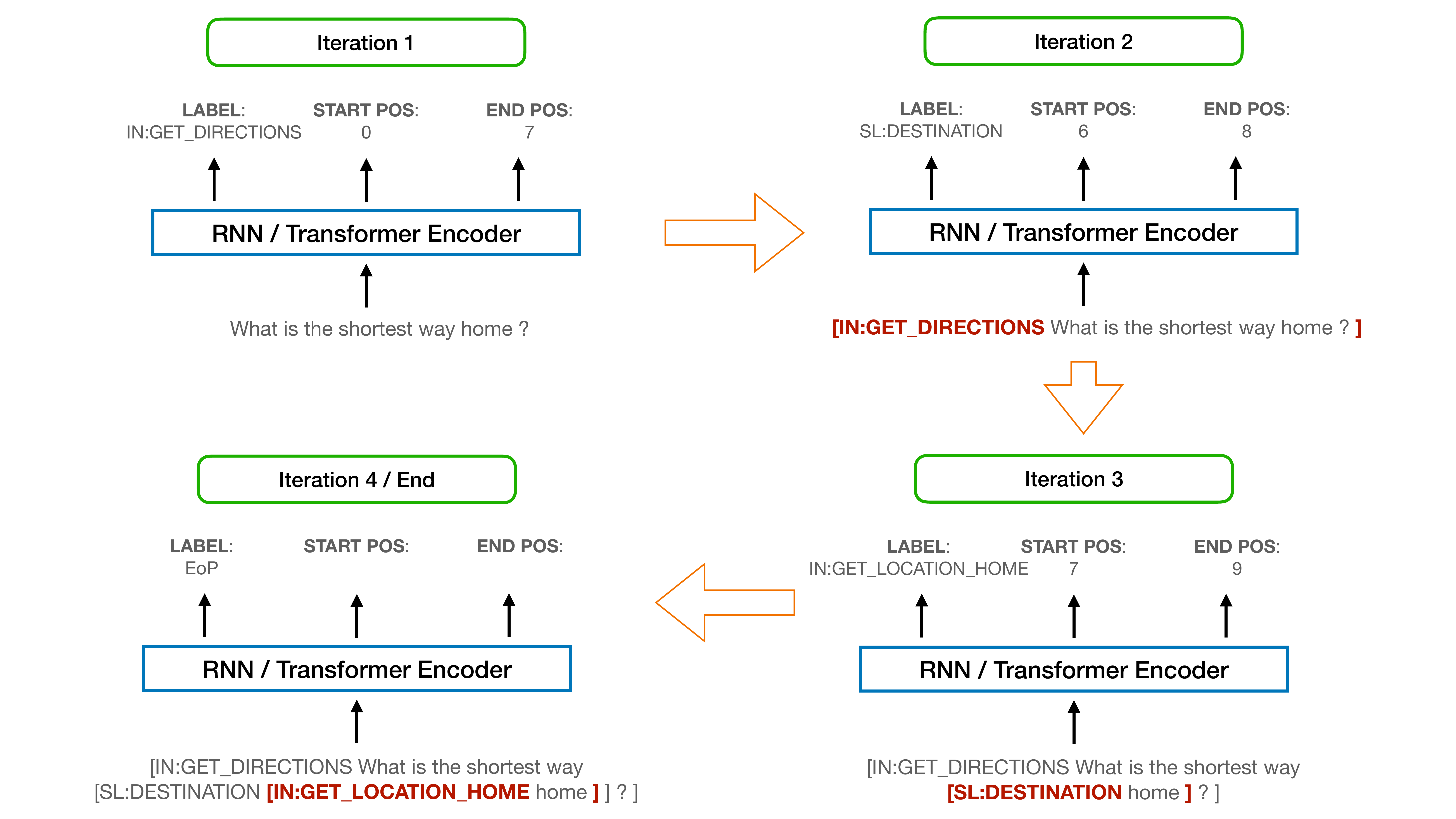}
    \caption{Overview of the top-down generation of the semantic parse tree corresponding to the utterance \textit{What is the shortest way home?} from the {\tt TOP} dataset \citep{Gupta2018SemanticPF} using our proposed model. The inserted labels at each generation step are highlighted in \textcolor{red}{red}.}
    \label{fig:gen_overview}
\end{figure*}

While we focus on semantic parsing in task-oriented dialog, we demonstrate that our model design is applicable to other structured prediction tasks, such as nested named entity recognition (nested NER). We empirically show that our model with no specific tuning performs on par with state-of-the-art machine reading comprehension approach for nested NER \citep{Li2020AUM} that was explicitly designed for that task.

\iffalse\alert{measure decoding speed improvement if any and mention that proposed approach has less parameters than seq2seq model?}\fi

% Yi's comments
% Diagrams how the approach works. Diagram for both RINE & insertion Transformer 
% Recursive part add it (example mathematical formula). How does the T-1 step influence the generation of step T
% Extended more add diagram. This is sentence then this is output structure. Compare to diagram of Don't Parse Insert (without decoder) 
% Formulation of model & 

\section{Proposed Approach}
%In this section we introduce the RINE model for conversational semantic parsing. We start by describing the process of transforming the semantic parse data for our model in Section ~\ref{subsec:io}. We then explain the training and inference procedure of our model in Sections ~\ref{subsec:train} and ~\ref{subsec:infer} respectively. 

%The semantic parse tree provides a structured description of tasks supported by dialogue agent and is similar to the design of constituency syntax tree.
First, we introduce the problem of semantic parsing in task-oriented dialog and give a general description of our approach in Section~\ref{sec:overview}. Then in Section~\ref{sec:training}, we give a detailed description of the forward pass and loss calculation in our model. Finally in Section~\ref{sec:generation} we describe the generation procedure of semantic parse tree given the input utterance.

\subsection{Overview} \label{sec:overview}

Given the utterance $X=(x_0,...,x_{n-1})$ with $n$ tokens, our goal is to predict the semantic parse tree $Y$\iffalse as shown in Figure \alert{TODO}\fi. Each leaf node in the tree $Y$ corresponds to a token $x_i \in X$, while each non-terminal node covers some span $(i,j)$ with tokens $x_{i:j} = (x_i, ..., x_{j-1})$. The label $l$ of each non-terminal node is either an intent (prefixed with {\tt IN:}) or a slot (prefixed with {\tt SL:}). The root node of the {\tt TOP} tree covering the span $(0, n)$ must be an intent. Intents can be nested inside the slots and vice versa resulting in the composite tree structures. It should be noted that this formulation of semantic parse tree resembles the constituent structures commonly adopted in syntactic parsing. The key difference is that the non-terminal nodes in {\tt TOP} tree are semantic entities in a dialog frame representation (i.e. intents and slots). Therefore they are not syntactic units and their corresponding leaf sub-sequences do not necessarily pass the constituency test. Instead, these non-terminal semantic nodes \emph{govern} the linguistic expressions where the meaning is derived: slot nodes dominate over the string spans denoting their values; intent nodes dominate over both the span of utterance signaling the intent and slot nodes as arguments of each intent.

For our approach, it is helpful to view the target tree $Y$ as the result of the incremental insertions of the elements in the set $S=\{(l_1, i_1, j_1), ..., (l_T, i_T, j_T)\}$ into the utterance $X$. The $t^{\text{th}}$ element $(l_t, i_t, j_t)$ in the set $S$ consists of the intent/slot label $l_t$, the start position $i_t$ and the end position $j_t$. The label $l_t$ covers the span $(i_t, j_{t}-1)$ in the partially build tree $Y_{t-1}$. The result of the sequence of consecutive insertions of all elements in set $S$ is the target tree $Y_T=Y$. The utterance $X$ is used as the input for the first insertion step.

You can see the example of the semantic tree generation using our model in Figure \ref{fig:gen_overview}. At the first iteration, the label {\tt IN:GET\_DIRECTIONS} is inserted at the start position $0$ and end position $7$ into the utterance \textit{What is the shortest way home ?}. The result of the insertion operation is the tree \textit{{\tt [IN:GET\_DIRECTIONS} What is the shortest way home ? {\tt ]}}. This tree is fed back into the model to output the tuple ({\tt SL:DESTINATION}, $6$, $8$). The updated tree with inserted label {\tt SL:DESTINATION} is fed back into the model to output the ({\tt IN:GET\_LOCATION\_HOME}, $7$, $9$). Finally, the model predicts a special end of prediction {\tt EoP} label that indicates the termination of the generation process.
%At generation step $t$, the label $l_t$ is inserted at the positions $i_t$ and $j_t$ into the linearized TOP tree $Y_{t-1}$ from the previous step $t-1$.

%The main idea behind our approach is to build the semantic parse tree by incrementally inserting the non-terminals (intent/slot labels) into the utterance. Figure \alert{TODO} shows an example of how we can build such semantic parse tree for the utterance \textit{What is causing traffic this afternoon}. To get a semantic parse tree, we first insert the intent {\tt IN:GET\_INFO\_TRAFFIC} as the root node of the semantic tree. We then insert the slot {\tt SL:DATE\_TIME} as the branch node with span \textit{this afternoon} linked under it.

%Since we operate 
%The insertion operation of non-terminal node (intent/slot) into the parse tree is equivalent to insertion of intent/slot label at the equivalent start and end positions in the linearized version of parse tree. The start and end positions represent the opening and closing of the non-terminal in the linearized tree. In the previous example, inserting {\tt IN:GET\_INFO\_TRAFFIC} as the root in the tree corresponds to inserting symbol {\tt IN:GET\_INFO\_TRAFFIC} at the first and last positions in the utterance. Similarly inserting {\tt SL:DATE\_TIME} as the branch node in the next step, corresponds to inserting this slot at the fifth and last positions in the linearized parse tree \alert{fix error}.

\subsection{Training} \label{sec:training}

For an input sequence $[w_1, w_2, ..., w_m]$ consisting of the tokens from utterance $X$ and intent/slot labels from parse tree $Y$, our model first encodes the input into a sequence of hidden vectors $[e_1, e_2, ..., e_m]$. This model consists of an encoder that can have an RNN \citep{Elman1990FindingSI}, Transformer \citep{vaswani2017attention} or any other architecture. In practice, we use the pretrained Transformer model RoBERTa \citep{Liu2019RoBERTaAR} due to its significant improvement in performance across natural language understanding tasks \citep{Wang2018GLUEAM} and state-of-the-art performance on task-oriented semantic parsing \citep{Rongali2020DontPG}.  

The hidden vector $e_1$ corresponding to the special start of the sentence symbol is passed to a multilayer perceptron (MLP) to predict the probability of the output label $l_t = \text{softmax}(\text{MLP}(e_1))$. % We use the attention probabilities from separate heads of the last multi-head self-attention layer \citep{vaswani2017attention} to predict the start and end positions $i_t$ and $j_t$. 
We use the attention probabilities from the first attention head of the last layer’s multi-head attention layer to predict the begin position $i_t$. Similarly, we use the attention probabilities from the second attention head of the last layer's multi-head attention layer to predict the end position $j_t$. This choice allows the model to extrapolate to start and end positions larger than the ones encountered during training. 

After getting the outputs from the model, we train it by combining three objectives: \textit{label loss} $\mathcal{L}_{\text{label}} = -\log p (l^{*}_{t} | Y^{*}_{t-1})$, \textit{start position loss} $\mathcal{L}_{\text{start}} = -\log p (i^{*}_{t} | Y^{*}_{t-1})$ and \textit{end position loss} $\mathcal{L}_{\text{end}} = -\log p (j^{*}_{t} | Y^{*}_{t-1})$. As a result, we minimize the joint negative log likelihood of the the ground-truth labels $(l^{*}_{t}, i^{*}_{t}, j^{*}_{t})$ given the ground-truth partial tree $Y^{*}_{t-1}$:
\begin{align*}
\mathcal{L} = \mathcal{L}_{\text{label}} + \mathcal{L}_{\text{start}} + \mathcal{L}_{\text{end}}
\end{align*}

During training, we batch the predictions across the generation time-steps $t=1,...,T$. We follow a top-down generation ordering to create the training set of pairs of partially constructed ground-truth trees $Y^{*}_{t-1}$ and outputs $(l^{*}_t, i^{*}_t, j^{*}_t))$. When using top-down ordering, we first generate the root node and then work down the tree to generate remaining nodes. In principle, we can use other generation orderings (such as bottom-up ordering where we start from generating nodes at the lowest level of the tree and work up, which we empirically compare against in Section~\ref{subsec:genorder}). 

%We feed the partially constructed linearized semantic parse tree $X$ into the model to predict the triplet $(i, j, l)$, where $l$ is the intent/slot label, $i$ and $j$ are the start and end positions of the label in the sentence. \alert{talk about how we obtain the partially constructred trees}

%The input $X$ is represented as the sequence of tokens which are transformed into the continuous representation by the model using multi-head self-attention and feedforward layers. The MLP classifier predicts the intent/slot label given the representation of the final layer. We use the attention map from the self-attention layer to predict the positions. Attention map allows us to generalize beyond the positions seen during training. We use the attention map obtained from different heads in self-attention to predict start and end positions respectively. 

\subsection{Generation} \label{sec:generation}

\begin{figure*}
    \centering
    \adjustbox{trim={.0\width} {.28\height} {0.0\width} {.28\height},clip}{\includegraphics[width=0.99\textwidth,height=0.99\textheight,keepaspectratio]{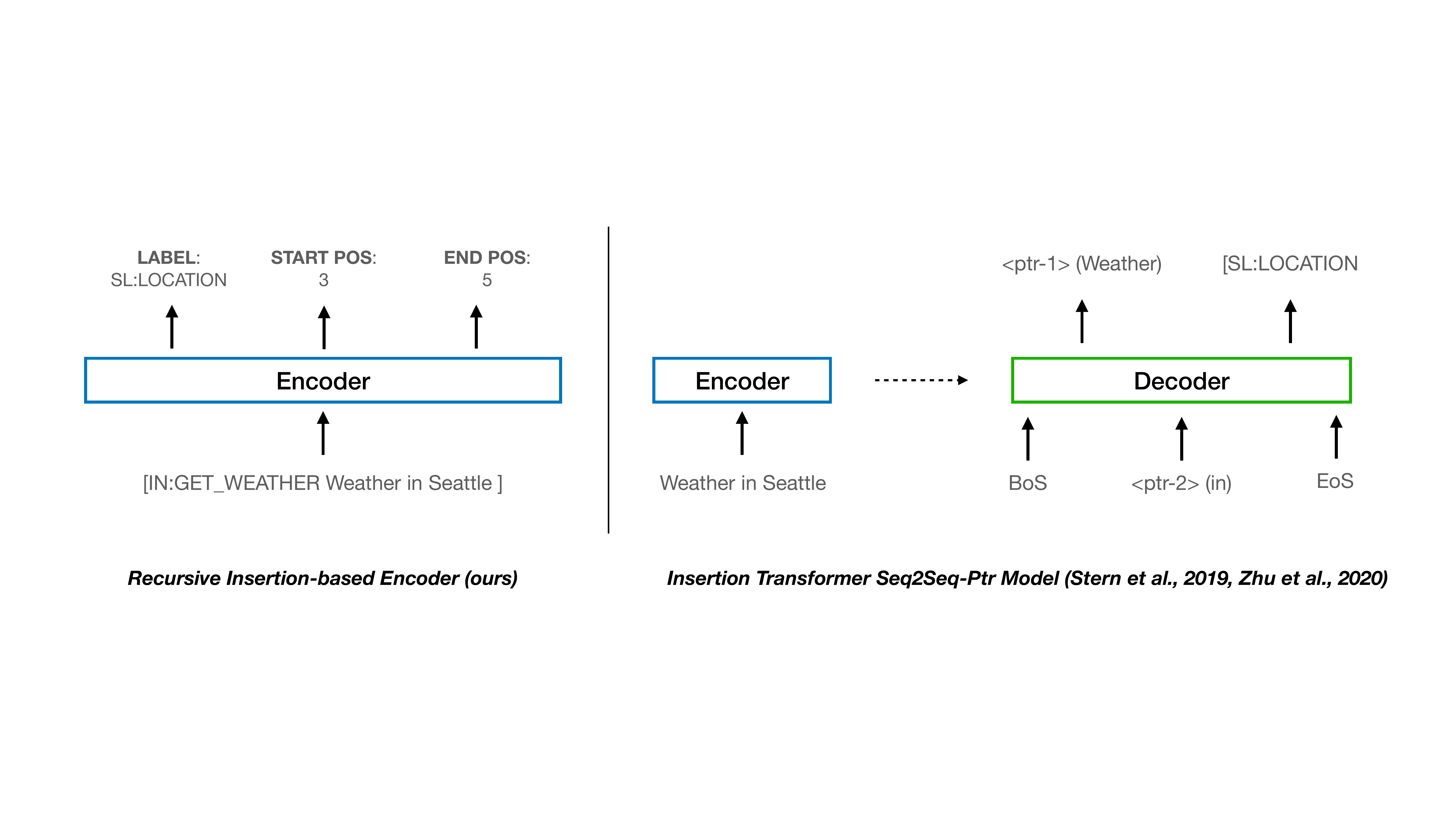}}
    \caption{Side-by-side comparison of two closely related architectures for semantic parsing in task-oriented dialog. On the left, we show the forward pass of our model. On the right, we show the forward pass of Insertion Transformer Seq2seq-Ptr Model \citep{Stern2019InsertionTF, Zhu2020DontPI}. Tokens \textit{$<$ ptr-i$>$} denote the pointers to the utterance. The Insertion Transformer follows the balanced binary tree ordering by predicting labels for each insertion slot (there are $T-1$ insertion slots for sentence of length $T$). Our model follows the top-down generation ordering by predicting the single intent/slot label with start and end positions in the linearized tree.} 
    \label{fig:comparison}
\end{figure*}

When evaluating the trained model, we use greedy decoding. We start from the input utterance $X$ and predict the most likely label $\hat{l}_1 = \arg \max_{l} p(l_{1} | Y_{0})$ and most likely start and end positions $\hat{i}_1 = \arg \max_{i} p(i_{1} | Y_{0})$ and $\hat{j}_1 = \arg \max_{j} p(j_{1} | Y_{0})$. We then insert the label $\hat{l}_1$ into position $\hat{i}_1$, followed by inserting the closing bracket into the position $\hat{j}_1 + 1$. We feed the resulting tree $\hat{Y}_1$ back into the model to predict the next triplet of ($\hat{l}_2$, $\hat{i}_2$, $\hat{j}_2$) following the same procedure. The entire process is repeated until the special end of prediction symbol {\tt EoP} is predicted as the most likely label by the model. %This generation procedure approximately finds the most likely semantic parse tree $\hat{Y}$ corresponding to input $X$.

%The generation procedure consists of local decisions made by the model at each timestep. At the generation step $t$, we feed the result of insertion operation from previous timestep $\hat{Y}_{t-1}$ into the model to predict the following tuple of $(l_t, i_t, j_t)$.

%Once the model is trained, we carry out the decoding procedure by iteratively building the semantic parse tree. Starting from the utterance, the model predicts the triplet of intent/slot label, start position and end position. The predicted label is inserted in the utterance at the predicted start and end positions. This modified utterance which is a partially constructed linearized parse tree is fed into the model again to predict the next triplet of intent/slot label, start position and end position. We repeat this procedure until the special end-of-sentence symbol \textit{eos} is predicted as the label by the model. This greedy decoding procedure is similar to the greedy decoding used in the sequence to sequence models \citep{Sutskever2014SequenceTS}. Unlike seq2seq model our approach doesn't use a separate decoder network to carry out the generation. \alert{show figure}

%\subsection{Discussion}

% Highlight differences between this model and don't parse generate/insert earlier in intro

\section{Related Work}

Parsing the meaning of utterances in task-oriented dialogue has been a prevalent problem in a research community since the advent of the {\tt ATIS} dataset \citep{Hemphill1990TheAS}. Traditionally, this task is formulated as a joint intent classification and slot tagging problem. Sequence labeling models based on recurrent neural networks (RNNs) \citep{Mesnil2013InvestigationOR,Liu2016AttentionBasedRN} and pre-trained Transformer models \citep{Devlin2019BERTPO,Chen2019BERTFJ} have been successfully used for joint intent classification and slot tagging. These sequence models can only parse \textit{flat} utterances which contain a single intent class and single slot label per each token in the utterance. To deal with this limitation, recent studies investigated structured prediction models based on neural shift-reduce parsers \citep{Dyer2016RecurrentNN,Gupta2018SemanticPF,Einolghozati2019ImprovingSP}, neural span-based parsers \citep{Stern2017AMS,Pasupat2019SpanbasedHS}, autoregressive sequence-to-sequence models \citep{Rongali2020DontPG,Aghajanyan2020ConversationalSP}, and non-autoregressive sequence-to-sequence models \citep{Zhu2020DontPI,Shrivastava2021SpanPN,Babu2021NonAutoregressiveSP} for handling \textit{compositional} queries. All of these approaches have been adapted from constituency parsing, dependency parsing and machine translation. Among these, the approach for task-oriented dialogue by \citet{Zhu2020DontPI} bears the most similarity to our model. 

%Among these, two approaches for task-oriented dialogue by \citet{Pasupat2019SpanbasedHS,Zhu2020DontPI} bear the most similarity to our model. 

%\citet{Pasupat2019SpanbasedHS} adapted a neural span-based semantic parser \citep{Stern2017AMS} for task-oriented dialogue. Similar to our model, their approach only contains the encoder network. Unlike the approach by \citep{Pasupat2019SpanbasedHS}, our model does not \textit{independently} predict labels for each span and does not use the CKY decoding algorithm \citep{Kasami1965AnER,Younger1967RecognitionAP} for generating the valid parse tree. 

\citet{Zhu2020DontPI} adapted the Insertion Transformer \citep{Stern2019InsertionTF} into the seq2seq-ptr model \citep{Rongali2020DontPG} for conversational semantic parsing. The  Insertion Transformer generates the linearized parse tree in balanced binary tree order by predicting labels in the insertion slots at each generation step (there are $T-1$ insertion slots for sentence of length $T$). Unlike traditional seq2seq models which scale linearly with length of target sequence, the Insertion Transformer only requires a logarithmic number of decoding steps. Despite sharing the insertion operation, there are several key differences between our approach and the Insertion Transformer seq2seq-ptr approach illustrated in Figure~\ref{fig:comparison}. Unlike Insertion Transformer: 1) our model does not have separate decoder, 2) our model generates a parse tree in a top-down fashion with the number of decoding steps equivalent to the number of intent/slot labels in the tree. Additionally, the termination strategy in our model is as simple as termination in vanilla seq2seq models. To terminate generation the Insertion Transformer requires predicting {\tt EoS} token for each insertion slot. This makes the {\tt EoS} token more frequent than other tokens which leads to generation of short target sequences. To avoid this issue, \citet{Zhu2020DontPI} add a special penalty hyperparameter to control the sequence length.

In parallel to the design of neural architectures, there has been a research effort on improving neural conversational semantic parsers in low-resource setting using meta-learning \citep{Chen2020LowResourceDA} and label semantics \citep{Athiwaratkun2020AugmentedNL,Paolini2021StructuredPA,Desai2021LowResourceTS}. These approaches are architecture agnostic and can be easily combined with our model to further improve performance.

%In parallel to the design of neural architectures, there has been a research effort on improving performance of semantic parsers for task-oriented dialogue with limited training data available. \citet{Chen2020LowResourceDA} used meta-learning for quick adaption of seq2seq conversational semantic parsers to new domains that contain a small amount of training data. \citet{Athiwaratkun2020AugmentedNL,Paolini2021StructuredPA,Desai2021LowResourceTS} incorporated label semantics that treat intent/slot labels as natural language outputs instead of separate discrete entities in seq2seq models to improve performance in few-slot and low-resource settings. These approaches are architecture agnostic and can be easily combined with our model to further improve performance in low-resource settings in task-oriented dialog. 

% add hierarchical ner or amr parsing (some version of dependency parsing) will add quite a bit of value to the paper.
% is it just this dataset? why not evaluated on other tasks/datasets

% say something about ordering (does it matter or not?). it might be task-dependant?

% revision of predictions. revision of slot labels given intent and vice-versa. (part of future work) not what we experimented yet. anchored vs non-anchored generation (required of representation). structure is properly nested tree. handling so-called non-continous (discontinuent) constituent parsing or non-projective dependencies. 

% hierarchical gets more improvement. verify that ablation.

% non-continuous constituents  non-projective dependencies unanchored meaning representations

\section{Experiments}

\iffalse
\begin{table*}[t!]
\centering
\input{tables/top-stats.tex}
\label{tab:topstats}
\end{table*}
\fi

\subsection{Datasets}

We use the {\tt TOP} \citep{Gupta2018SemanticPF} and {\tt TOPv2} \citep{Chen2020LowResourceDA} conversational semantic parsing datasets as well as {\tt ACE2005} nested named entity recognition dataset in our experiments. 

The {\tt TOP} dataset\footnote{\url{http://fb.me/semanticparsingdialog}} \citep{Gupta2018SemanticPF} consists of natural language utterances in two domains: \textit{navigation} and \textit{event}. The dataset consists of $25$ intents and $36$ slots. Following previous work \citep{Einolghozati2019ImprovingSP,Rongali2020DontPG,Zhu2020DontPI}, we remove the utterances that contain the \texttt{UNSUPPORTED} intent from the dataset. This results in $28{,}414$ train, $4{,}032$ valid and $8{,}241$ test utterances\footnote{The dataset statistics were verified with authors of \citep{Zhu2020DontPI}}. $39\%$ of queries in the {\tt TOP} dataset are hierarchical.

The {\tt TOPv2} dataset \citep{Chen2020LowResourceDA} is an extension of {\tt TOP} dataset \citep{Gupta2018SemanticPF} that was collected by following the same guidelines. Following the experimental setup of \citet{Chen2020LowResourceDA} we use low-resource versions of \textit{reminder} and \textit{weather} domains. The \textit{reminder} domain consists of $19$ intents and $32$ slots. $21\%$ of queries in \textit{reminder} domain are hierarchical. The \textit{weather} domain consists of $7$ intents and $11$ slots. All queries in the \textit{weather} domain are flat. The low-resource data was created by taking a fixed number of training \textit{samples per intent and slot label (SPIS)} from the original dataset. If a particular intent or slot occurred less than the specified number of times then all the parse trees containing that intent or slot are selected. We use the same train, validation and test data at $25$ and $500$ SPIS for \textit{reminder} and \textit{weather} prepared by \citet{Chen2020LowResourceDA}\footnote{\url{https://fb.me/TOPv2Dataset}}. The \textit{reminder} domain at $500$ SPIS contains $4{,}788$ train and $2{,}526$ valid samples, \textit{weather} $500$ SPIS contains $2{,}372$ train and $2{,}667$ valid samples, \textit{reminder} $25$ SPIS contains $493$ train and $337$ valid samples, and \textit{weather} $25$ SPIS contains $176$ train and $147$ valid samples. For both SPIS settings the test splits of \textit{reminder} and \textit{weather} contain $5{,}767$ and $5{,}682$ test samples respectively.

The {\tt ACE2005} nested named entity recognition dataset is derived from the {\tt ACE2005} corpus \citep{Walker2006ACE} and consists of sentences from a variety of domains, including news and online forums. We use the same processing and splits of \citet{Li2020AUM}, resulting in $7,299$ sentences for training, $971$ for validation, and $1,060$ for testing. The dataset has seven entity types: \textit{location, organization, person, vehicle, geographical entity, weapon, facility}. $38\%$ of queries in the {\tt ACE2005} dataset are hierarchical. The design of the semantic parse trees in {\tt ACE2005} dataset is similar to the design of semantic parse trees in {\tt TOP} dataset. The entities in the {\tt ACE2005} dataset are represented as slots. Slots in {\tt ACE2005} can be nested inside the slots resulting in nested entity structures.

Following previous work \citep{Gupta2018SemanticPF,Einolghozati2019ImprovingSP,Rongali2020DontPG,Zhu2020DontPI,Aghajanyan2020ConversationalSP} we use exact match (EM) accuracy as the metric for evaluating approaches on {\tt TOP} and {\tt TOPv2} datasets. The exact match measures the number of utterances where complete trees are correctly predicted by the model. On {\tt ACE2005} we report the span-level micro-averaged precision, recall and F1 scores.

\subsection{Hyperparameters}

We follow the experimental settings of previous conversational semantic parsing work \citep{Rongali2020DontPG,Zhu2020DontPI,Chen2020LowResourceDA} and use a pre-trained RoBERTa \citep{Liu2019RoBERTaAR} model as the backbone of our RINE model. We experiment with both \robertabase and \robertalarge architectures. The {\tt [CLS]} representation is passed into the MLP with $1$ hidden layer to predict the intent/slot label. The probabilities from the second and third heads of the last self-attention layer are used to predict start and end positions.

We train a sequence-to-sequence pointer network (seq2seq-ptr) that combines a Transformer \citep{vaswani2017attention} and pointer-generator network \citep{See2017GetTT}. \citet{Rongali2020DontPG} proposed this model for the task of conversational semantic parsing. The seq2seq-ptr model generates the linearized semantic parse tree by alternating between generating intent/slot tags from a fixed vocabulary and copying a token from the source query using a pointer network \cite{Vinyals2015PointerN,See2017GetTT}. The encoder of seq2seq-ptr is initialized using a pre-trained RoBERTa \citep{Liu2019RoBERTaAR} architecture. The decoder is initialized with random weights. The decoder contains $6$ layers, $4$ attention heads, $512$-dimensional embeddings, and $1{,}024$ hidden units.

\begin{table*}[t!]
\centering
\small
\centering
\begin{tabular}[t]{c|c|c}
\toprule
Method & Pretrained model & Exact Match \\
\midrule
RNNG \citep{Einolghozati2019ImprovingSP} & - & 80.86 \\
RNNG \citep{Einolghozati2019ImprovingSP} & ELMo & 86.26 \\
RNNG ensemble + SVMRank \citep{Einolghozati2019ImprovingSP} & ELMo & 87.25 \\
\midrule
Non-AR Seq2seq-Ptr \citep{Shrivastava2021SpanPN} & \robertabase & 85.07 \\
Seq2seq-Ptr \citep{Rongali2020DontPG} & \robertabase & 86.67 \\
Insertion Transformer + Seq2seq-Ptr \citep{Zhu2020DontPI} & \robertabase & 86.74 \\
Decoupled Seq2seq-Ptr \citep{Aghajanyan2020ConversationalSP} & \bartlarge & 87.10 \\
%\midrule
%Seq2seq-Ptr (our replication) & \robertabase & 85.73$\pm$0.21 \\
%Seq2seq-Ptr (our replication) & \robertalarge & 86.67$\pm$0.21 \\
\midrule
RINE (ours) & \robertabase & 87.14$\pm$0.06 \\
RINE (ours) & \robertalarge & \textbf{87.57$\pm$0.03} \\
\bottomrule
\end{tabular}
\caption{Accuracy (exact match $\uparrow$) on the test split of {\tt TOP} dataset \citep{Gupta2018SemanticPF}. Non-AR stands for non-autoregressive. Pretrained model stands for the type of pretrained architecture used in the corresponding method.}
\label{tab:topresults}
\end{table*}

\begin{table*}[t!]
\centering
\small
\centering
\begin{tabular}[t]{c|c|cccc}
\toprule
\multirow{3}{*}{Method} & \multirow{3}{*}{Pretrained model} & \multicolumn{4}{c}{Exact Match} \\
& & \multicolumn{2}{c}{Reminder} & \multicolumn{2}{c}{Weather} \\
& & 25 SPIS & 500 SPIS & 25 SPIS & 500 SPIS \\
\midrule
LSTM Seq2Seq-Ptr \citep{Chen2020LowResourceDA} & - & 21.5 & 65.9 & 46.2 & 78.6 \\ 
Seq2seq-Ptr \citep{Chen2020LowResourceDA} & \robertabase & - & 71.9 & - & 83.5 \\
Seq2seq-Ptr \citep{Chen2020LowResourceDA} & \bartlarge & 55.7 & 71.9 & 71.6 & 84.9 \\
%\midrule
%Seq2Seq-Ptr (our replication) & \robertabase & 60.33$\pm$2.09 & 78.46$\pm$0.3 & 57.98$\pm$2.02 & 85.41$\pm$0.04 \\
%Seq2Seq-Ptr (our replication) & \robertalarge & - \iffalse 11.77$\pm$1.78 \fi & 79.32$\pm$0.78 & - \iffalse 2.31$\pm$0.55 \fi & 86.62$\pm$0.13 \\
\midrule
RINE (ours) & \robertabase & 68.71$\pm$0.46 & 80.30$\pm$0.04 & 74.53$\pm$0.86 & \textbf{87.80$\pm$0.04} \\
RINE (ours) & \robertalarge & \textbf{71.10$\pm$0.63} & \textbf{81.31$\pm$0.22} & \textbf{77.03$\pm$0.16} & 87.50$\pm$0.28 \\
\bottomrule
\end{tabular}
\caption{Accuracy (exact match $\uparrow$) on the test split of \textit{reminder} and \textit{weather} domains of {\tt TOPv2} dataset \citep{Chen2020LowResourceDA}. SPIS stands for \textit{samples for each intent and slot label}.}
\label{tab:topv2results}
\end{table*}

We use the Adam optimizer \citep{kingma2014adam} with the following hyperparameters: $\beta_1=0.9$, $\beta_2=0.98$, $\epsilon=1e-6$ and $L_2$ weight decay of $1e-4$. When using \robertabase, we warm-up the learning rate for $500$ steps up to a peak value of $5e-4$ and then decay it based on the inverse square root of the update number. When using \robertalarge, we warm-up the learning rate for $1{,}000$ steps up to a peak value of $1e-5$ and then decay it based on the inverse number of update steps. The same hyperparameters of the optimizer are used for training both RINE and seq2seq models. We use a dropout \citep{Srivastava2014DropoutAS} rate of $0.3$ and an attention dropout rate of $0.1$ in both our proposed models and seq2seq baseline. The choices of hyperparameters were made based on preliminary experiments on {\tt TOP}. For all datasets we use $4$ Tesla V100 GPUs to train both baseline and proposed model. We use $3$ random seeds to train all models and report the average and standard deviation. For fair comparison of results, we train both baseline and proposed model for the same number of iterations on all datasets. We implement all models on top of the \textit{fairseq} framework \citep{ott2019fairseq}. 

\subsection{Results}

\begin{table*}[t!]
\centering
\small
\centering
\begin{tabular}[t]{c|c|ccc}
\toprule
Method & Pretrained Model & Precision & Recall & F1 \\
\midrule
Hyper-Graph LSTM \citep{Katiyar2018NestedNE} & - & 70.6 & 70.4 & 70.5 \\
Seg-Graph \citep{Wang2018NeuralSH} & GLoVE & 76.8 & 72.3 & 74.5 \\
ARN \citep{Lin2019SequencetoNuggetsNE} & GLoVE & 76.2 & 73.6 & 74.9 \\
Path-BERT \citep{Shibuya2019NestedNE} & \bertlarge & 82.98 & 82.42 & 82.7 \\
Merge-BERT \citep{Fisher2019MergeAL} & \bertlarge & 82.7 & 82.1 & 82.4 \\
DYGIE \citep{Luan2019AGF} & GLoVE + ELMo & - & - & 82.9 \\
Seq2seq-BERT \citep{Strakov2019NeuralAF} & \bertlarge & - & - & 84.33 \\
TANL \citep{Paolini2021StructuredPA} & \tfivebase & - & - & 84.9 \\
BERT-MRC \citep{Li2020AUM} & \bertlarge & \textbf{87.16} & 86.59 & \textbf{86.88} \\
\midrule
RINE (ours) & \robertabase & 84.13$\pm$0.03 & 87.06$\pm$0.19 & 85.57$\pm$0.1 \\
RINE (ours) & \robertalarge & 84.62$\pm$0.05 & \textbf{88.33$\pm$0.07} & 86.44$\pm$0.04 \\
\bottomrule
\end{tabular}
\caption{Precision ($\uparrow$), recall ($\uparrow$) and F1 score ($\uparrow$) on the test split of {\tt ACE2005} dataset.}
\label{tab:ace2005results}
\end{table*}

\subsubsection{TOP dataset}
We present the results on {\tt TOP}~\citep{Gupta2018SemanticPF} in Table \ref{tab:topresults}. Our proposed RINE model that uses \robertalarge as the backbone outperforms all previously published results on the {\tt TOP} dataset. In particular, our RINE model initialized with the \robertalarge outperforms the non-autoregressive seq2seq-ptr model \citep{Shrivastava2021SpanPN} initialized with \robertabase by $2.5$ EM, autoregressive seq2seq-ptr models \citep{Rongali2020DontPG,Zhu2020DontPI} initialized with \robertabase by $0.9$ EM, decoupled seq2seq-ptr model initialized with \bartlarge by $0.47$ EM, and RNNG ensemble with SVM reranking \citep{Einolghozati2019ImprovingSP} by $0.32$ EM. Our model initialized with the \robertabase outperforms the non-autoregressive seq2seq-ptr model \citep{Shrivastava2021SpanPN} initialized with \robertabase by $2.0$ EM,  autoregressive seq2seq-ptr models \citep{Rongali2020DontPG,Zhu2020DontPI} initialized with \robertabase by $0.4$ EM, and performs on par with RNNG ensemble with SVM reranking \citep{Einolghozati2019ImprovingSP} and decoupled seq2seq-ptr \citep{Aghajanyan2020ConversationalSP} initialized with \bartlarge. Unlike RNNG, we do not use ensembling and do not rerank outputs of our model. Unlike decoupled seq2seq-ptr, we don't use stochastic weight averaging \citep{Izmailov2018AveragingWL} to improve results. 

We also re-implemented and trained our seq2seq-ptr model by \citet{Rongali2020DontPG}. We obtain exact match of $85.73\pm0.21$ and $86.67\pm0.21$ with \robertabase and \robertalarge backbones respectively.
Compared to our replication of autoregressive seq2seq-ptr model by \citet{Rongali2020DontPG}, the proposed RINE model achieves $1.41$ and $0.9$ improvement in exact match using \robertabase and \robertalarge respectively. The performance of RINE is $5-7$ standard deviations higher than performance of our seq2seq-ptr model.

The decomposition of the parse tree leads to multiple training passes over the same source sentence in our model. This is unlike seq2seq-ptr model that processes each pair of source and target once during the epoch. One could argue that increasing the number of training iterations in baseline seq2seq-ptr approach can lead to the same performance as our model. However despite training baseline seq2seq-ptr for the same iterations as RINE, seq2seq-ptr stops improving validation exact match after $150$ epochs and starts overfitting after.
 
\subsubsection{TOPv2 dataset}
We present the results on low-resource versions of \textit{reminder} and \textit{weather} domains in {\tt TOPv2}~\citep{Chen2020LowResourceDA} in Table \ref{tab:topv2results}. There are several observations on this dataset. 

% remove word significantly
First, our proposed RINE model outperforms the baseline autoregressive seq2seq-ptr model on all evaluated scenarios of this dataset. In the $500$ SPIS setting with \robertabase, our RINE model achieves $8.4$ and $2.9$ exact match improvement on \textit{reminder} and \textit{weather} domains over the best published seq2seq-ptr baseline. In the $25$ SPIS setting with \robertabase, our RINE model achieves $13.0$ and $2.9$ exact match improvement on \textit{reminder} and \textit{weather} domains over the best published autoregressive seq2seq-ptr baseline. In the \textit{reminder} domain the improvement in performance is higher for the $25$ SPIS setting, whereas in the \textit{weather} domain the improvement in performance is comparable for both $25$ and $500$ SPIS. We hypothesize that this is due to the \textit{reminder} domain being more challenging that \textit{weather} domain since it contains composite utterances and have a larger number of intent and slot types.

Second, we trained our re-implementation of seq2seq-ptr model by \citet{Rongali2020DontPG} on $500$ SPIS setting. We find that our replication of seq2seq-ptr approach with \robertabase backbone achieves $78.46\pm0.3$ and $85.41\pm0.04$ exact match on \textit{reminder} and \textit{weather} domains. The performance of our replication of seq2seq-ptr model with \robertabase performs better than seq2seq-ptr with \robertabase reported by \citet{Chen2020LowResourceDA}. We believe it is due to the number of epochs we used to train seq2seq-ptr models. \citet{Chen2020LowResourceDA} report using 100 epochs to train all models without meta-learning, whereas we train seq2seq-ptr for larger number of epochs to match the number of iterations used to train RINE. Despite training our seq2seq-ptr for larger number of iterations, RINE outperforms our replication of seq2seq-ptr approach by $1.84$ and $2.39$ exact match on $500$ SPIS setting of the \textit{reminder} and \textit{weather} domains respectively. We make a similar observation to {\tt TOP} dataset and find that seq2seq-ptr approach underperforms RINE despite increasing number of training iterations to match the number of training iterations of RINE. In particular, seq2seq-ptr converges to the highest validation exact match after $300$ epochs and starts overfitting after.

\iffalse
Second, we find that our proposed model scales better with larger encoder architectures compared to the seq2seq-ptr model. In particular, in the $25$ SPIS setting the seq2seq-ptr model with the \robertalarge encoder underperforms the same model  with the \robertabase encoder with difference of up to $56$ EM. Whereas when using our model, we consistently see a $1-2.5$ exact match improvement when increasing the size of encoder except for $500$ SPIS \textit{weather} domain where both architectures perform comparably.
\fi

%Finally, our proposed approach performs on par with the seq2seq-ptr approach that utilizes optimization-based meta-learning \citep{Nichol2018OnFM} with additional training data to improve performance \citep{Chen2020LowResourceDA}. This shows that the magnitude of improvement obtained from better modeling choices can be as significant as utilizing additional training data with meta-learning.

\subsubsection{ACE2005}
We present the results on {\tt ACE2005} dataset in Table~\ref{tab:ace2005results}. Our model outperforms all previously published approaches designed for nested named entity recognition task, except for BERT-MRC \citep{Li2020AUM} approach. In particular, RINE with \robertalarge encoder underperforms BERT-MRC with \bertlarge encoder in terms of precision, while achieving higher recall and comparable F1 score.
Compared to our model, BERT-MRC uses questions constructed from the annotation guideline notes used for collecting ACE2005 dataset. These questions contain the ground-truth label semantics (example for ORG label the question is ”find organizations including companies, agencies and institutions“). \citet{Li2020AUM} show that label semantics improve results by 1.5 F1 (Table 5 of BERT-MRC \citep{Li2020AUM}) and use scores obtained with label semantics in the main results. We believe using some form of label semantics in the output can further improve RINE on ACE2005. Despite the lack of label semantics in our approach and no additional tuning, RINE achieves comparable performance to state-of-the-art BERT-MRC which further demonstrates the strong performance of our model.

\section{Analysis}

\subsection{Does generation order matter?} \label{subsec:genorder}

In this section we analyze whether generation order matters for our model. In our default setting, we follow the top-down ordering of the labels when training and generating trees. We experiment with a bottom-up ordering and present the comparison with top-down ordering in Table~\ref{tab:genorder}. We find that the choice of ordering makes no difference in exact match accuracy on the validation split of {\tt TOP} suggesting that the model is agnostic to the particular order in which it was trained.

\subsection{Flat vs composite queries}
% valid tree structure?
Our initial motivation of the proposed approach stems from the argument that the seq2seq-ptr models are not ideally suited for parsing the utterance into hierarchically structured representations. In this section, we empirically validate this argument by breaking down the performance of both models on flat and hierarchical trees. We find that a larger improvement of our approach over the baseline model comes from composite trees. In particular, on the validation split of {\tt TOP}, the RINE model achieves $3.4\%$ relative improvement on the composite queries over the seq2seq-ptr model. On the same dataset, the RINE model achieves $1.6\%$ improvement on the flat queries over the seq2seq-ptr model. We make a similar empirical observation on the validation split of the \textit{reminder} domain on the {\tt TOPv2} dataset. In the $25$ SPIS setting, the proposed model achieves a $22.4\%$ relative improvement on composite queries, while it achieves only a $6\%$ relative improvement on flat queries. In the $500$ SPIS setting, the proposed model achieves a $3.5\%$ and $1.3\%$ relative improvement on composite and flat queries. The relative improvement in performance on composite queries becomes larger in the low-resource $25$ SPIS setting. This shows that the proposed approach is better suited for hierarchically structured meaning representations compared to the seq2seq-ptr model.

\subsection{Validity of generated trees}

\begin{table}[t!]
\centering
\small
\centering
\begin{tabular}[t!]{cc|c}
\toprule
Architecture & Order & Exact Match \\
\midrule
\multirow{2}{*}{\robertabase} & Top-down & 87.07$\pm$0.09 \\
 & Bottom-up & 87.01$\pm$0.04 \\
\midrule
\multirow{2}{*}{\robertalarge} & Top-down & 87.72$\pm$0.07 \\
 & Bottom-up & 87.70$\pm$0.04 \\
\bottomrule
\end{tabular}
\caption{Exact match of proposed RINE model on validation split of {\tt TOP} dataset with top-down and bottom-up generation orderings.}
\label{tab:genorder}
\end{table}

In this section we compare the validity of the semantic parse trees generated by our and baseline approaches. Unlike our approach which generates perfectly valid trees when trained on both low- and high-resource settings of {\tt TOP} dataset, seq2seq-ptr struggles to achieve perfect validity when trained in the low-resource setting. In particular in the $25$ SPIS setting of the \textit{reminder} domain, $93\%$ of the generated trees by seq2seq-ptr model are valid. When we increase the training dataset size and train the seq2seq-ptr model on the $500$ SPIS setting of the \textit{reminder} domain, the validity of generated trees  becomes close to $100\%$. This demonstrates that baseline seq2seq-ptr model requires larger amount of training data to learn the structure of semantic parse trees in order to generate valid trees. 

\subsection{Generation Efficiency}

In this section we compare the generation latency of the seq2seq-ptr and RINE approaches. We generate parse trees by processing 1 sentence at a time using Tesla V100 GPU. As the measure of generation efficiency we use number of sentences per second ($\uparrow$) processed by each approach. We show results in Table~\ref{tab:efficiency}. We find that our approach is $3.5\times$ faster with \robertabase and $2\times$ faster with \robertalarge than seq2seq-ptr approach. We notice that the decoding efficiency of our approach relative to baseline drops when using large encoder. Seq2seq-ptr scales better with larger encoders due to caching of the encoder representations that are later used by the decoder.

\section{Conclusions and Future Work}

\begin{table}[t!]
\centering
\small
\centering
\begin{tabular}{c|cc}
\toprule
Architecture & Seq2seq-ptr & RINE \\
\midrule
\robertabase & 3.70 & \textbf{12.95} \\
\robertalarge & 3.42 & \textbf{7.09} \\
\bottomrule
\end{tabular}
\caption{Generation efficiency (sentences per second $\uparrow$) of seq2seq-ptr and RINE approaches with \robertabase and \robertalarge architectures on validation split of {\tt TOP} \citep{Gupta2018SemanticPF} dataset.}
\label{tab:efficiency}
\end{table}

Following on the exciting and recent development of hierarchically structured meaning representations for task-oriented dialog, we proposed a recursive insertion-based encoder approach that achieves state-of-the-art results on low- and high-resource versions of the conversational semantic parsing benchmark {\tt TOP} \citep{Gupta2018SemanticPF,Chen2020LowResourceDA}. \iffalse Starting from the input query, our approach incrementally builds the semantic parse tree by inserting the non-terminal intent/slot labels into the input.\fi We also showed that the proposed model design is applicable to nested NER, where it achieves comparable results to state-of-the-art with no additional tuning. Analysis of the results demonstrates that proposed approach achieves higher relative improvement on hierarchical trees compared to baselines and does not require large amount of training data to learn the structure of the trees. 

Despite achieving strong empirical results, there are some limitations with the  proposed approach. The insertion-based approach is generally limited to generate \emph{anchored} and well-nested tree structures that are addition to the input with no deletion or reordering. While such assumption is commonly adopted in many linguistic representations, well-known exceptions do exist (e.g. non-projective dependencies~\citep{hall:2008}, discontiguous constituents~\citep{vijay-shanker:1987,mueller:2004}, or unanchored meaning representations such as AMR~\citep{banarescu:2013}). While the authors of {\tt TOP} \citep{Gupta2018SemanticPF} found that a very small fraction of English queries ($0.3\%$) require a more general unanchored graph-based meaning representation, we believe it is important to address such issues for non-configurational languages as well as the broader range of application-specific structural representations. We plan to extend the model with additional actions such as token insertion and swap in order to support parsing of such non-anchored representation in multilingual semantic parsing. 

Overall, we hope that our paper inspires research community to further extend and apply our model for a variety of structured prediction tasks.
\\ \\
\small{\textbf{Acknowledgements}:
We would like to thank Daniele Bonadiman, Arshit Gupta, James Gung, Yassine Benajiba, Haidar Khan, Saleh Soltan and other members of Amazon AWS AI for helpful suggestions.}

\newpage
\bibliography{main}

\newpage

\appendix
\section{Qualitative Comparison} \label{sec:appendix}
We show several examples comparing generated parse trees of seq2seq-ptr model and RINE in Table~\ref{tab:qualitative}.

\begin{table*}[t!]
\small
\centering
\begin{tabular}{p{2.25cm}p{13cm}}
\toprule
Type & Output \\
\midrule
Source & summer concerts \\
Target & {\tt [IN:GET\_EVENT [SL:DATE\_TIME} summer {\tt SL:DATE\_TIME] [SL:CATEGORY\_EVENT} concerts {\tt SL:CATEGORY\_EVENT] IN:GET\_EVENT]} \\
Seq2seq-ptr & {\tt [IN:GET\_EVENT [SL:CATEGORY\_EVENT} summer concerts {\tt SL:CATEGORY\_EVENT] IN:GET\_EVENT]} \\
RINE & {\tt [IN:GET\_EVENT [SL:DATE\_TIME} summer {\tt SL:DATE\_TIME] [SL:CATEGORY\_EVENT} concerts {\tt SL:CATEGORY\_EVENT] IN:GET\_EVENT]} \\
\midrule
Source & Art parties for grownups in Nashville \\
Target & {\tt [IN:GET\_EVENT [SL:CATEGORY\_EVENT} Art parties {\tt SL:CATEGORY\_EVENT]} for {\tt [SL:ATTRIBUTE\_EVENT} grownups {\tt SL:ATTRIBUTE\_EVENT]} in {\tt [SL:LOCATION} Nashville {\tt SL:LOCATION] IN:GET\_EVENT]} \\
Seq2seq-ptr & {\tt [IN:GET\_EVENT [SL:CATEGORY\_EVENT} Art parties for grownups {\tt SL:CATEGORY\_EVENT]} in {\tt [SL:LOCATION} Nashville {\tt SL:LOCATION] IN:GET\_EVENT]} \\
RINE & {\tt [IN:GET\_EVENT [SL:CATEGORY\_EVENT} Art parties {\tt SL:CATEGORY\_EVENT]} for {\tt [SL:ATTRIBUTE\_EVENT} grownups {\tt SL:ATTRIBUTE\_EVENT]} in {\tt [SL:LOCATION} Nashville {\tt SL:LOCATION] IN:GET\_EVENT]} \\
\midrule
Source & is their a ton of traffic near balboa \\
Target & {\tt [IN:GET\_INFO\_TRAFFIC} is their a ton of traffic {\tt [SL:LOCATION [IN:GET\_LOCATION [SL:SEARCH\_RADIUS} near {\tt SL:SEARCH\_RADIUS] [SL:LOCATION} balboa {\tt SL:LOCATION] IN:GET\_LOCATION] SL:LOCATION] IN:GET\_INFO\_TRAFFIC]} \\
Seq2seq-ptr & {\tt [IN:GET\_INFO\_TRAFFIC} is their a {\tt [SL:DATE\_TIME} ton of {\tt SL:DATE\_TIME]} traffic {\tt [SL:LOCATION [IN:GET\_LOCATION [SL:SEARCH\_RADIUS} near {\tt SL:SEARCH\_RADIUS] [SL:LOCATION} balboa {\tt SL:LOCATION] IN:GET\_LOCATION] SL:LOCATION] IN:GET\_INFO\_TRAFFIC]} \\
RINE & {\tt [IN:GET\_INFO\_TRAFFIC} is their a ton of traffic {\tt [SL:LOCATION [IN:GET\_LOCATION [SL:SEARCH\_RADIUS} near {\tt SL:SEARCH\_RADIUS] [SL:LOCATION} balboa {\tt SL:LOCATION] IN:GET\_LOCATION] SL:LOCATION] IN:GET\_INFO\_TRAFFIC]} \\
\midrule
Source & Directions to Thriving Minds from Aiden 's school , need to arrive by 4 pm . \\
Target & {\tt [IN:GET\_DIRECTIONS} Directions to {\tt [SL:DESTINATION [IN:GET\_LOCATION [SL:POINT\_ON\_MAP} Thriving Minds {\tt SL:POINT\_ON\_MAP] IN:GET\_LOCATION] SL:DESTINATION]} from {\tt [SL:SOURCE [IN:GET\_LOCATION\_SCHOOL [SL:CONTACT} Aiden {\tt SL:CONTACT]} 's school {\tt IN:GET\_LOCATION\_SCHOOL] SL:SOURCE]} , need to arrive {\tt [SL:DATE\_TIME\_ARRIVAL by 4 pm SL:DATE\_TIME\_ARRIVAL]} . {\tt IN:GET\_DIRECTIONS]} \\
Seq2seq-ptr & {\tt [IN:GET\_DIRECTIONS} Directions to {\tt [SL:DESTINATION [IN:GET\_LOCATION\_SCHOOL [SL:CONTACT} Thriving {\tt SL:CONTACT]} Minds {\tt IN:GET\_LOCATION\_SCHOOL] SL:DESTINATION]} from {\tt [SL:SOURCE [IN:GET\_LOCATION\_SCHOOL [SL:CONTACT} Aiden {\tt SL:CONTACT]} 's school {\tt IN:GET\_LOCATION\_SCHOOL] SL:SOURCE]} , need to arrive {\tt [SL:DATE\_TIME\_ARRIVAL} by 4 pm {\tt SL:DATE\_TIME\_ARRIVAL] . IN:GET\_DIRECTIONS]} \\
RINE & {\tt [IN:GET\_DIRECTIONS} Directions to {\tt [SL:DESTINATION [IN:GET\_LOCATION [SL:POINT\_ON\_MAP} Thriving Minds {\tt SL:POINT\_ON\_MAP] IN:GET\_LOCATION] SL:DESTINATION]} from {\tt [SL:SOURCE [IN:GET\_LOCATION\_SCHOOL [SL:CONTACT} Aiden {\tt SL:CONTACT]} 's school {\tt IN:GET\_LOCATION\_SCHOOL] SL:SOURCE]} , need to arrive {\tt [SL:DATE\_TIME\_ARRIVAL by 4 pm SL:DATE\_TIME\_ARRIVAL]} . {\tt IN:GET\_DIRECTIONS]} \\
\bottomrule
\end{tabular}
\caption{Qualitative comparison of generated trees by seq2seq and RINE approaches on the validation split of {\tt TOP} dataset. We show examples where RINE generates correct parse tree and seq2seq-ptr generates incorrect parse tree.}

%Canonical & \texttt{[IN:SEND\_MESSAGE [SL:CONTENT [} I'll be there at 6pm \texttt{] ] ]} \\
%[IN:GET_EVENT [SL:DATE_TIME summer SL:DATE_TIME] [SL:CATEGORY_EVENT concerts SL:CATEGORY_EVENT] IN:GET_EVENT]
\label{tab:qualitative}
\end{table*}

% sentences per sec
% RINE (RoBERTa base): 12.95 (sentences/sec)
% RINE (RoBERTa large): 7.09  (sentences/sec)
% Seq2seq-ptr (RoBERTa base):  3.61 (sentences/sec)
% Seq2seq-ptr (RoBERTa large): ???  (sentences/sec)

\end{document}